\documentclass{article} 
\usepackage[accepted]{icml2020}

\usepackage{multirow}
\usepackage{subcaption}
\usepackage{hyperref}
\usepackage{url}
\usepackage{graphicx}
\usepackage{amsmath}
\usepackage{bbm}

\newcommand{\mb}{\mathbf}

\newcommand{\mc}{\mathcal}

\newtheorem{definition}{\textsc{Definition}}

\newcommand{\our}{\textsc{Seg-Bert}}

\newcommand{\gbert}{\textsc{Graph-Bert}}
\newcommand{\bert}{\textsc{Bert}}
\newcommand{\transformer}{\textsc{Transformer}}

\begin{document}

\twocolumn[
\icmltitle{Segmented {\gbert} for Graph Instance Modeling}



\icmlsetsymbol{equal}{*}

\begin{icmlauthorlist}
\icmlauthor{Jiawei Zhang}{ifmlab}
\end{icmlauthorlist}

\icmlaffiliation{ifmlab}{IFM Lab, Department of Computation, Florida State University, Tallahassee, FL, USA.}

\icmlcorrespondingauthor{Jiawei Zhang}{jiawei@ifmlab.org}

\icmlkeywords{Machine Learning, ICML}

\vskip 0.3in
]

\printAffiliationsAndNotice{}

\begin{abstract}

In graph instance representation learning, both the diverse graph instance sizes and the graph node orderless property have been the major obstacles that render existing representation learning models fail to work. In this paper, we will examine the effectiveness of {\gbert} on graph instance representation learning, which was designed for node representation learning tasks originally. To adapt {\gbert} to the new problem settings, we re-design it with a segmented architecture instead, which is also named as {\our} (Segmented {\gbert}) for reference simplicity in this paper. {\our} involves no node-order-variant inputs or functional components anymore, and it can handle the graph node orderless property naturally. What's more, {\our} has a segmented architecture and introduces three different strategies to unify the graph instance sizes, i.e., \textit{full-input}, \textit{padding/pruning} and \textit{segment shifting}, respectively. {\our} is pre-trainable in an unsupervised manner, which can be further transferred to new tasks directly or with necessary fine-tuning. We have tested the effectiveness of {\our} with experiments on seven graph instance benchmark datasets, and {\our} can out-perform the comparison methods on six out of them with significant performance advantages.

\end{abstract}
\section{Introduction}\label{sec:introduction}

Different from the previous graph neural network (GNN) works \cite{zhang2020graph,Velickovic_Graph_ICLR_18,Kipf_Semi_CORR_16}, which mainly focus on the node embeddings in large-sized graph data, we will study the representation learning of the whole graph instances in this paper. Representative examples of graph instance data studied in research include the human brain graph, molecular graph, and real-estate community graph, whose nodes (usually only in tens or hundreds) represent the brain regions, atoms and POIs, respectively. Graph instance representation learning has been demonstrated to be an extremely difficult task. Besides the diverse input graph instance sizes, attributes and extensive connections, the inherent node-orderless property \cite{Meng_Isomorphic_NIPS_19} brings about more challenges on the model design. 

Existing graph representation learning approaches are mainly based on the convolutional operator \cite{NIPS2012_4824} or the approximated graph convolutional operator \cite{Hammond_2011,NIPS2016_6081} for pattern extraction and information aggregation. By adapting CNN \cite{NIPS2012_4824} or GCN \cite{Kipf_Semi_CORR_16} to graph instance learning settings, \cite{Niepert_Learning_16,verma2018graph,Zhang2018AnED,xinyi2018capsule,Chen_Dual_19} introduce different strategies to handle the node orderless properties. However, due to the inherent learning problems with graph convolutional operator, such models have also been criticized for serious performance degradation on deep architectures \cite{zhang2019gresnet}. Deep sub-graph pattern learning \cite{Meng_Isomorphic_NIPS_19} is another emerging new-trend on graph instance representation learning. Different from these aforementioned methods, \cite{Meng_Isomorphic_NIPS_19} introduces the IsoNN (Isomorphic Neural Net) to learn the sub-graph patterns automatically for graph instance representation learning, which requires the identical-sized graph instance input and cannot handle node attributes.

In this paper, we introduce a new graph neural network, i.e., {\our} (\underline{Se}gmented \textsc{\underline{G}raph-Bert}), for graph instance representation learning based on the recent {\gbert} model. Originally, {\gbert} is introduced for node representation learning \cite{zhang2020graph}, whose learning results have been demonstrated to be effective for various node classification/clustering tasks already. Meanwhile, to adapt {\gbert} to the new problem settings, we re-design it in several major aspects: (a) {\our} involves no node-order-variant inputs or functional components. {\our} excludes the nodes' relative positional embeddings from the initial inputs; whereas both the self-attention \cite{Vaswani_Attention_17} and the representation fusion component in {\our} can both handle the graph node orderless property naturally. (b) {\our} unifies graph instance input to fit the model configuration. {\our} has a segmented architecture and introduces three strategies to unify the graph instance sizes, i.e., \textit{full-input}, \textit{padding/pruning} and \textit{segment shifting}, which help feed {\our} with the whole input graph instances (or graph segments) for representation learning. (c) {\our} re-introduces graph links as initial inputs. Considering that some graph instances may have no node attributes, {\our} re-introduces the graph links (in an artificial fixed node order) back as one part of the initial inputs for graph instance representation learning. (d) {\our} uses the graph residual terms of the whole graph instance. Since we focus on learning the representations of the whole graph instances, the graph residual terms involved in {\our} will be defined for all the nodes in the graph (or segments) instead of merely for the target nodes \cite{zhang2020graph}.

{\our} is pre-trainable in an unsupervised manner, and the pre-trained model can be further transferred to new tasks directly or with necessary fine-tuning. To be more specific, in this paper, we will explore the pre-training and fine-tuning of {\our} for several graph instance studies, e.g., \textit{node attribute reconstruction}, \textit{graph structure recovery}, and \textit{graph instance classification}. Such explorations will help construct the functional model pipelines for graph instance learning, and avoid the unnecessary redundant learning efforts and resource consumptions.

We summarize our contributions of this paper as follows:
\begin{itemize}

\item \textbf{Graph Representation Learning Unification}: We examine the effectiveness of {\gbert} in graph instance representation learning task in this paper. The success of this paper will help unify the currently disconnected representation learning tasks on nodes and graph instances, as discussed in \cite{zhang2019graph}, with a shared methodology, which will even allow the future model transfer across graph datasets with totally different properties, e.g., from social networks to brain graphs.

\item \textbf{Node-Orderless Graph Instances}: We introduce a new graph neural network model, i.e., {\our}, for graph instance representation learning by re-designing {\gbert}. {\our} only relies on the attention learning mechanisms and the node-order-invariant inputs and functional components in {\our} allow it to handle the graph instance node orderless properties very well.

\item \textbf{Segmented Architecture}: We design {\our} in a segmented architecture, which has a reasonable-sized input portal. To unify the diverse sizes of input graph instances to fit the input portals, {\our} introduces three strategies, i.e., \textit{full-input}, \textit{padding/pruning} and \textit{segment shifting}, which can work well for different learning scenarios, respectively.

\item \textbf{Pre-Train \& Transfer \& Fine-Tune}: We study the unsupervised pre-training of {\our} on graph instance studies, and explore to transfer such pre-trained models to the down-stream application tasks directly or with necessary fine-tuning. In this paper, we will pre-train {\our} with unsupervised \textit{node attribute reconstruction} and \textit{graph structure recovery} tasks, and further fine-tune {\our} on supervised \textit{graph classification} as the down-stream tasks.

\end{itemize}

The remaining parts of this paper are organized as follows. We will introduce the related work in Section~\ref{sec:related_work}. Detailed information about the {\our} model will be introduced in Section~\ref{sec:method}, whereas the pre-training and fine-tuning of {\our} will be introduced in Section~\ref{sec:analysis} in detail. The effectiveness of {\our} will be tested in Section~\ref{sec:experiment}. Finally, we will conclude this paper in Section~\ref{sec:conclusion}.

\section{Related Work}\label{sec:related_work}

Several interesting research topics are related to this paper, which include \textit{graph neural networks}, \textit{graph representation learning} and \textit{{\bert}}. 

\noindent \textbf{GNNs and Graph Representation Learning}: Different from the node representation learning \cite{Kipf_Semi_CORR_16,Velickovic_Graph_ICLR_18}, GNNs proposed for the graph representation learning aim at learning the representation for the entire graph instead \cite{Narayanan_Graph_17}. To handle the graph node permutation invariant challenge, solutions based various techniques, e.g., attention \cite{Chen_Dual_19,Meltzer_Permutation_19}, pooling \cite{Meltzer_Permutation_19,ranjan2019asap,Jiang_Gaussian_18}, capsule net \cite{Mallea_Capsule_19}, Weisfeiler-Lehman kernel \cite{NIPS2016_6166} and sub-graph pattern learning and matching \cite{Meng_Isomorphic_NIPS_19}, have been proposed. 

For instance, \cite{Mallea_Capsule_19} studies the graph classification task with a capsule network; \cite{Chen_Dual_19} defines a dual attention mechanism for improving graph convolutional network on graph representation learning; and \cite{Meltzer_Permutation_19} introduces an end-to-end learning model for graph representation learning based on an attention pooling mechanism. On the other hand, \cite{ranjan2019asap} focuses on studying the sparse and differentiable pooling method to be adopted with graph convolutional network for graph representation learning; \cite{NIPS2016_6166} examines the optimal assignments of kernels for graph classification, which can out-perform the Weisfeiler-Lehman kernel on benchmark datasets; and \cite{Jiang_Gaussian_18} proposes to introduce the Gaussian mixture model into the graph neural network for representation learning. As an emerging new-trend, \cite{Meng_Isomorphic_NIPS_19} explores the deep sub-graph pattern learning and proposes to learn interpretable graph representations by involving sub-graph matching into a graph neural network.

\noindent \textbf{{\bert}}: {\transformer} \cite{Vaswani_Attention_17} and {\bert} \cite{Bert} based models have almost dominated NLP and related research areas in recent years due to their great representation learning power. Prior to that, the main-stream sequence transduction models in NLP are mostly based on complex recurrent \cite{Hochreiter_Long_Neural_97,DBLP:journals/corr/ChungGCB14} or convolutional neural networks \cite{kim-2014-convolutional}. However, as introduced in \cite{Vaswani_Attention_17}, the inherently sequential nature precludes parallelization within training examples. To address such a problem, a brand new representation learning model solely based on attention mechanisms, i.e., the {\transformer}, is introduced in \cite{Vaswani_Attention_17}, which dispense with recurrence and convolutions entirely. Based on {\transformer}, \cite{{Bert}} further introduces {\bert} for deep language understanding, which obtains new state-of-the-art results on eleven natural language processing tasks. By extending {\transformer} and {\bert}, many new {\bert} based models, e.g., T5 \cite{raffel2019exploring}, ERNIE \cite{Sun_ERNIE} and RoBERTa \cite{Liu_RoBERTa}, can even out-perform the human beings on almost all NLP benchmark datasets. Some extension trials of {\bert} on new areas have also been observed. In \cite{zhang2020graph}, the authors explore to extend {\bert} for graph representation learning, which discard the graph links and learns node representations merely based on the attention mechanism.

\section{Problem Formulation}\label{sec:formulate}

In this section, we will first introduce the notations used in this paper. After that, we will provide the definitions of several important terminologies, and then introduce the formal statement of the studied problem.

\subsection{Notations}

In the sequel of this paper, we will use the lower case letters (e.g., $x$) to represent scalars, lower case bold letters (e.g., $\mb{x}$) to denote column vectors, bold-face upper case letters (e.g., $\mb{X}$) to denote matrices, and upper case calligraphic letters (e.g., $\mathcal{X}$) to denote sets or high-order tensors. Given a matrix $\mb{X}$, we denote $\mb{X}(i,:)$ and $\mb{X}(:,j)$ as its $i_{th}$ row and $j_{th}$ column, respectively. The ($i_{th}$, $j_{th}$) entry of matrix $\mb{X}$ can be denoted as either $\mb{X}(i,j)$. We use $\mb{X}^\top$ and $\mb{x}^\top$ to represent the transpose of matrix $\mb{X}$ and vector $\mb{x}$. For vector $\mb{x}$, we represent its $L_p$-norm as $\left\| \mb{x} \right\|_p = (\sum_i |\mb{x}(i)|^p)^{\frac{1}{p}}$. The Frobenius-norm of matrix $\mb{X}$ is represented as $\left\| \mb{X} \right\|_F = (\sum_{i,j} |\mb{X}(i,j)|^2)^{\frac{1}{2}}$. The element-wise product of vectors $\mb{x}$ and $\mb{y}$ of the same dimension is represented as $\mb{x} \otimes \mb{y}$, whose concatenation is represented as $\mb{x} \sqcup \mb{y}$.

\subsection{Terminology Definitions}

Here, we will provide the definitions of several important terminologies used in this paper, which include \textit{graph instance} and \textit{graph instance set}.

\begin{definition}
(Graph Instance): Formally, a graph instance studied in this paper can be denoted as $G=(\mc{V}, \mc{E}, w, x)$, where $\mc{V}$ and $\mc{E}$ denote the sets of nodes and links in the graph, respectively. Mapping $w: \mc{E} \to \mathbbm{R}$ projects links in the graph to their corresponding weight. For unweighted graphs, we will have $w(e_{i,j}) = 1, \forall e_{i,j} \in \mc{E}$ and $w(e_{i,j}) = 0, \forall e_{i,j} \in \mc{V} \times \mc{V} \setminus \mc{E}$. For the nodes in the graph instance, they may also be associated with certain attributes, which can be represented by mapping $x: \mc{V} \to \mc{X}$ (here, $\mc{X} = \mathbbm{R}^{d_x}$ denotes the attribute vector space and $d_x$ is the space dimension).
\end{definition}

Based on the above definition, given a node $v_i$ in graph instance $G$, we can represent its connection weights with all the other nodes in the graph as $\mb{w}_i = [w(e_{i,j})]_{v_j \in \mc{V}} \in \mathbbm{R}^{|\mc{V}| \times 1}$. Meanwhile, the raw attribute vector representation of $v_i$ can also be simplified as $\mb{x}_i = x(v_i)$. The size of graph instance $G$ can be denoted as the number of involved nodes, i.e., $|\mc{V}|$. For the graph instances studied in this paper, they can be in different sizes actually, which together can be represented as a \textit{graph instance set}.

\begin{definition}
(Graph Instance Set): For each graph instance studied in this paper, it can be attached with some pre-defined class labels. Formally, we can represent $n$ labeled graph instances studied in this paper as set $\mc{G} = \left\{(G_i, \mb{y}_i) \right\}_{i = 1}^n$, where $\mb{y}_i \in \mc{Y}$ denotes the label vector of $G_i$ (here, $\mc{Y} = \mathbbm{R}^{d_y}$ is the class label vector space and $d_y$ is the space dimension).
\end{definition}

For representation simplicity, in reference to the graph instance set (without labels), we can also denote it as $\mc{G}$, which will be used in the following problem statement.

\begin{figure*}[t]
    \begin{minipage}{\textwidth}
    \centering
    	\includegraphics[width=0.95\linewidth]{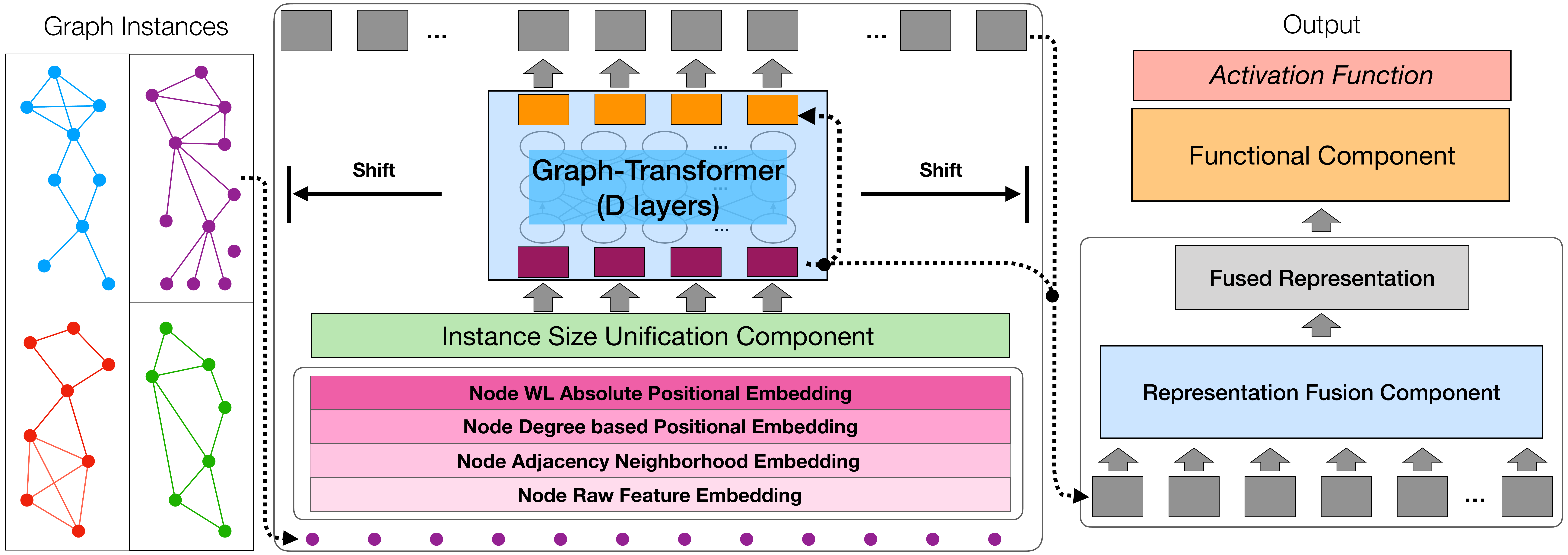}
    	\caption{An Illustration of the {\our} Model for Graph Instance Representation Learning.}
    	\label{fig:architecture}
    \end{minipage}%
    \vspace{-5pt}
\end{figure*}

\subsection{Problem Formulation}

Based on the notations and terminologies defined above, we can provide the problem statement as follows.

\noindent \textbf{Problem Statement}: Formally, given the labeled graph instance set $\mc{G}$, we aim at learning a mapping $f: \mc{G} \to \mc{R}^{d_h}$ to project the graph instances to their corresponding latent representations ($d_h$ denotes the hidden representation space dimension). What's more, we cast extra requirements on the mapping $f$ in this paper: (a) representations learned by $f$ should be invariant to node orders, (b) $f$ can accept graph instances in various sizes, as well as diverse categories of information inputs, (c) $f$ can be effectively pre-trained with the unsupervised learning tasks, and (d) representations learned by $f$ can also be transferred to the down-stream application tasks.

\section{The {\our} Model}\label{sec:method}

In this section, we will provide the detailed information about the {\our} model for graph instance representation learning. As illustrated in Figure~\ref{fig:architecture}, the {\our} model has several key components: (1) \textit{graph instance serialization} to reshape the various-sized input graph instances into node list, where the node orders will not affect the learning results; (2) \textit{initial embedding extraction} to define the initial input feature vectors for all nodes in the graph instance; (3) input size unification to fit the input portal size of \textit{graph-transformer}; (4) \textit{graph-transformer} to learn the nodes' representations in the graph instance (or segments) with several layers; (5) \textit{representation fusion} to integrate the learned representations of all nodes in the graph instance; and (6) \textit{functional component} to compute and output the learning results. These components will all be introduced in this section in detail, whereas the learning detail of {\our} will be discussed in the follow-up Section~\ref{sec:analysis} instead.


\subsection{Graph Serialization and Initial Embeddings}

Formally, given a graph instance $G \in \mc{G}$ from the graph set, we can denote its structure as $G = (\mc{V}, \mc{E}, w, x)$, involving node set $\mc{V}$ and link set $\mc{E}$, respectively. To simplify the notations, we will not indicate the graph instance index subscript in this section. Both the initial inputs and the functional components in {\our} are node-order-invariant, i.e., the nodes' learned representations are not dependent on the node orders. Therefore, regardless of the nodes' orders, we can serialize node set $\mc{V}$ into a sequence $[v_1, v_2, \cdots, v_{|\mc{V}|}]$. For the same graph instance, if it is fed to train/tune {\our} for multiple times, the node order in the list can change arbitrarily without affecting the learning representations.

For each node in the list, e.g., $v_i$, we can represent its raw features as a vector $\mb{x}_i = x(v_i) \in \mathbbm{R}^{d_x \times 1}$, which can cover various types of information, e.g., node tags, attributes, textual descriptions and even images. Via certain embedding mappings, we can denote the embedded feature representation of $v_i$'s raw features as
\begin{equation}
\mb{e}_i^{(x)} = \mbox{Embed} \left( \mb{x}_i \right) \in \mathbbm{R}^{d_h \times 1}.
\end{equation}
Depending on the input features, different approaches can be utilized to define the $\mbox{Embed}(\cdot)$ function, e.g., CNN for image features, LSTM for textual features, positional embedding for tags and MLP for real-number features. In the case when the graph instance nodes have no raw attributes on nodes, a dummy zero vector will be used to fill in the embedding vector $\mb{e}_i^{(x)}$ entries by default.

To handle the graph instances without node attributes, in this paper, we will also extend the original {\gbert} model by defining the node adjacency neighborhood embeddings. Meanwhile, to ensure such an embedding is node-order invariant, we will cast an artificially fixed node order on this embedding vector (which is fixed forever for the graph instance). For simplicity, we will just follow the node subscript index as the node orders in this paper. Formally, for each node $v_i$, following the fixed artificial node order, we can denote its adjacency neighbors as a vector $\mb{w}_i = \left[w(i,j)\right]_{v_j \in \mc{V}} \in \mathbbm{R}^{|\mc{V}| \times 1}$. Via several fully connected (FC) layers based mappings, we can represent the embedded representation of $v_i$'s adjacency neighborhood information as 
\begin{equation}
\mb{e}_i^{(w)} = \mbox{FC-Embed} \left( \mb{w}_i \right) \in \mathbbm{R}^{d_h \times 1}.
\end{equation}

The degrees of nodes can illustrate their basic properties \cite{Chung_Spectra_03,Bondy_Graph_76}, and according to \cite{Lovasz1996} for the Markov chain or random walk on graphs, their final stationary distribution will be proportional to the nodes' degrees. Node degree is also a node-order invariant actually. Formally, we can represent the degree of $v_i$ in the graph instance as $D(v_i) \in \mathbbm{N}$, and its embedding can be represented as 
\begin{equation}
\begin{aligned}
\mb{e}_i^{(d)} &= \mbox{Position-Embed}\left( \mbox{D}(v_i) \right)\\
&= \left[sin\left (\frac{\mbox{D}(v_i)}{10000^{\frac{2 l}{d_{h}}}} \right), cos\left(\frac{\mbox{D}(v_j)}{10000^{\frac{2 l + 1}{d_{h}}}} \right) \right]_{l=0}^{\left \lfloor \frac{d_h}{2} \right \rfloor},
\end{aligned}
\end{equation}
where $\mb{e}_i^{(d)} \in \mathbbm{R}^{d_h \times 1}$ and the vector index $l$ will iterate through the vector to compute the entry values based on the $sin(\cdot)$ and $cos(\cdot)$ functions.

In addition to the node raw feature embedding, node adjacency neighborhood embedding and node degree embedding, we will also include the nodes' Weisfeiler-Lehman role embedding vector in this paper, which effectively denotes the nodes' global roles in the input graph. Nodes' Weisfeiler-Lehman code is node-order-invariant, which denotes a positional property of the nodes actually. Formally, given a node $v_i$ in the input graph instance, we can denote its pre-computed WL code as $\mbox{WL}(v_i) \in \mathbbm{N}$, whose corresponding embeddings can be denoted as
\begin{equation}
\begin{aligned}
\mb{e}_i^{(r)} &= \mbox{Position-Embed}\left( \mbox{WL}(v_i) \right) \in \mathbbm{R}^{d_h \times 1}.
\end{aligned}
\end{equation}

{\our} doesn't include the \textit{relative positional embedding} and \textit{relative hop distance embedding} used in \cite{zhang2020graph}, as there exist no target node for the graph instances studied in this paper. Based on the above descriptions, we can represent the initially computed input embedding vectors of node $v_i$ in graph $G$ as 
\begin{equation}\label{equ:initial_embedding}
\mb{h}^{(0)}_i = \mbox{sum} \left( \mb{e}_i^{(x)}, \mb{e}_i^{(w)}, \mb{e}_i^{(d)}, \mb{e}_i^{(r)} \right) \in \mathbbm{R}^{d_h \times 1}.
\end{equation}


\subsection{Graph Instance Size Unification Strategies}

Different from \cite{zhang2020graph}, where the sampled sub-graphs all have the identical size, the graph instance input usually have different number of nodes instead. To handle such a problem, we design {\our} with an instance size unification component in this paper. Formally, we can denote the input portal size of {\our} used in this paper as $k$, i.e., it can take the initial input embedding vectors of $k$ nodes at a time. Depending on the input graph instance sizes, {\our} will define the parameter $k$ and handle the graph instances with different strategies:
\begin{itemize}
\item \textbf{Full-Input Strategy}: The input portal size $k$ of {\our} is defined as the largest graph instance size in the dataset, and dummy node padding will be used to expand all graph instances to $k$ nodes (zero padding for the connections, raw attributes and other tags).

\item \textbf{Padding/Pruning Strategy}: The input portal size $k$ of {\our} is assigned with a value slightly above the graph instance average size. For the graph instance input with less than $k$ nodes, dummy node padding (zero padding) is used to expand the graph to $k$ nodes; whereas for larger input graph instances, a $k$-node sub-graph (e.g., the first $k$ nodes) will be extracted from them and the remaining nodes will be pruned.

\item \textbf{Segment Shifting Strategy}: A fixed input portal size $k$ will be pre-specified, which can be a very small number. For the graph instance input with node set $\mc{V}$, the nodes will be divided into $\left \lceil \frac{|\mc{V}|}{k} \right \rceil$ segments, and dummy node padding will be used for the last segment if necessary. {\our} will shift along the segments to learn all the nodes representations in the graph instances.
\end{itemize}

Generally, the \textit{full-input strategy} will use all the input graph nodes for representation learning, but for a small-graph set with a few number of extremely large graph instance(s), a very large $k$ will be used in {\our}, which may introduce unnecessary high time costs. The \textit{padding/pruning strategy} balances the parameter $k$ among all the graph instances and can learn effective representations in an efficient way, but it may have information loss for the pruned parts of some graph instances. Meanwhile, the \textit{segment shifting strategy} can balance between the \textit{full-input strategy} and \textit{padding/pruning strategy}, which fuses the graph instance global information for representation learning with a small model input portal. More experimental tests of such different graph instance size unification strategies will also be explored with experiments on real-world benchmark datasets to be introduced in Section~\ref{sec:experiment}.



\begin{figure*}
    \centering
    \begin{subfigure}[b]{.23\textwidth}
    	\includegraphics[width=\linewidth]{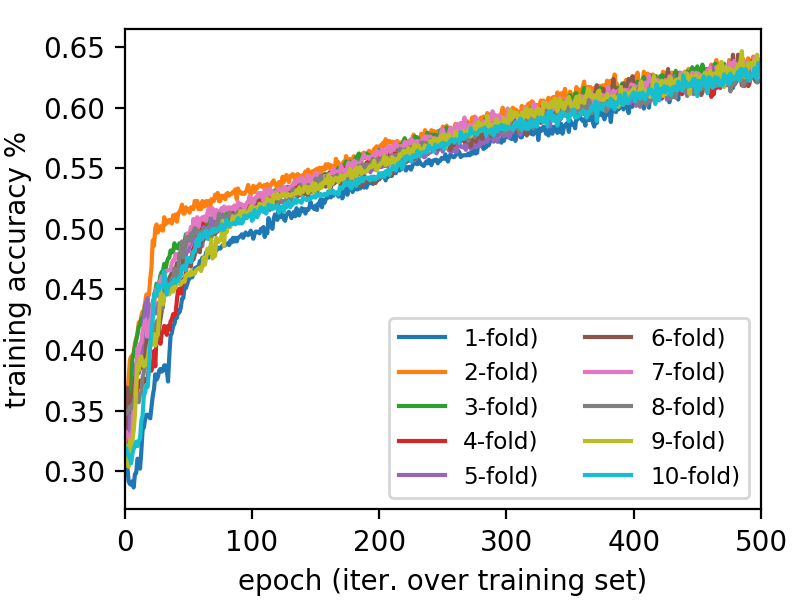}
    	\caption{Train Acc (IMDB)}\label{fig:acc_train}
    \end{subfigure}%
    \hfill
    \begin{subfigure}[b]{.23\textwidth}
    	\includegraphics[width=\linewidth]{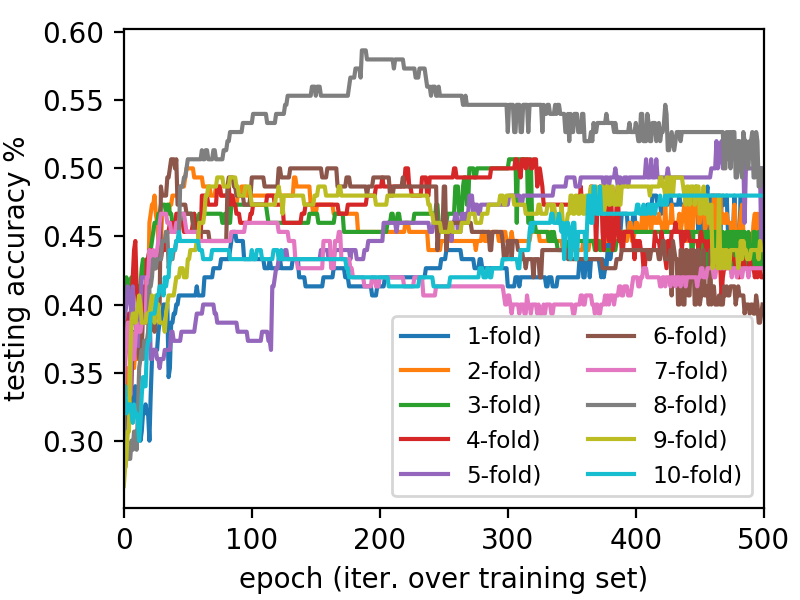}
    	\caption{Test Acc (IMDB)}\label{fig:acc_test}
    \end{subfigure}%
    \hfill
    \begin{subfigure}[b]{.23\textwidth}
    	\includegraphics[width=\linewidth]{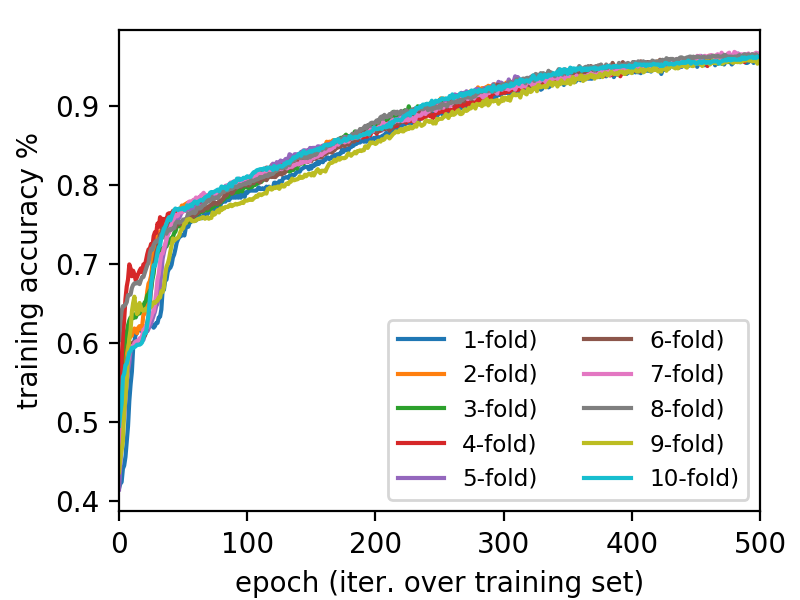}
    	\caption{Train Acc (Proteins)}\label{fig:acc_train}
    \end{subfigure}%
    \hfill
    \begin{subfigure}[b]{.23\textwidth}
    	\includegraphics[width=\linewidth]{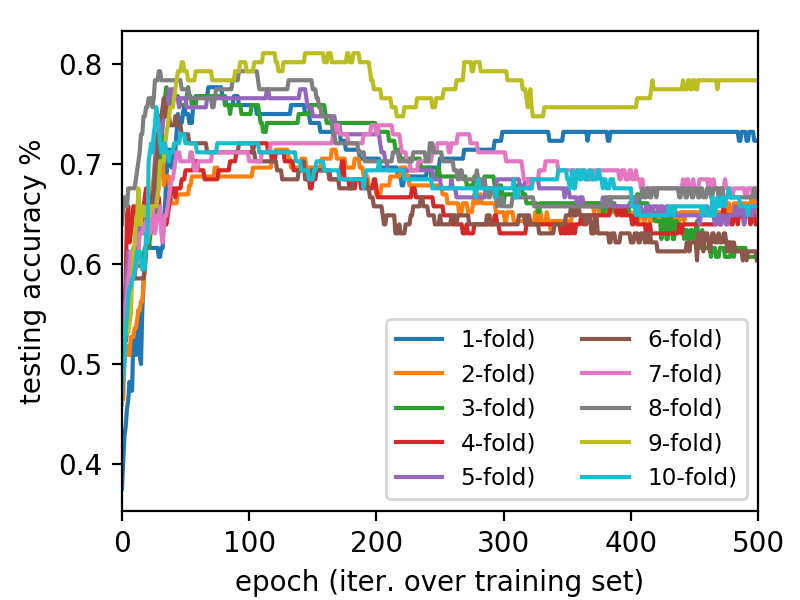}
    	\caption{Test Acc (Proteins)}\label{fig:acc_test}
    \end{subfigure}%
    \vspace{-5pt}
    \caption{Learning records of {\our} on the IMDB-Multi social-graph dataset and the Proteins bio-graph dataset. For the graph data with discrete structures, most of the comparison models studied in the experiments will overfit the training data easily and an early stop is usually necessary. The x axis: iteration, and the y axis: training/testing loss.}\label{fig:graph_bert_protein}
    \vspace{-10pt}
\end{figure*}

\subsection{Segmented Graph-Transformer}


To learn the graph instance representations, we will introduce the segmented graph-transformer in this part to update such nodes' representations iteratively with $D$ layers. Formally, based on the above descriptions, we can denote the current segment as $s_j = [v_{j,1}, v_{j,2}, \cdots, v_{j,k}]$, whose initial input embeddings can be denoted as $\mb{H}_j^{(0)} = [ \mb{h}^{(0)}_{j,1}, \mb{h}^{(0)}_{j,2}, \cdots, \mb{h}^{(0)}_{j,k} ]^\top \in \mathbbm{R}^{k \times d_h}$ and  $\mb{h}^{(0)}_{j,i} \in  \mathbbm{R}^{d_h \times 1}$ is defined in Equation~(\ref{equ:initial_embedding}). Depending on the different unification strategies adopted, the segment can denote all the graph nodes, pruned/padded node subset or the node segments, respectively. For the segmented graph-transformer, it will update the segment representations iteratively for each layer $\forall l \in \{1, \cdots, D\}$ according to the following equation:
\begin{equation}
\begin{aligned}
\mb{H}_j^{(l)} &= \mbox{G-Transformer} ( {\mb{H}}_j^{(l-1)} ),\\
&= \mbox{Transformer} ({\mb{H}}_j^{(l-1)}) + \mbox{G-Res} ({\mb{H}}_j^{(l-1)}, \mb{X}_j),
\end{aligned}
\end{equation}
where $\mbox{Transformer} (\cdot)$ and $\mbox{G-Res} (\cdot)$ denote the transformer \cite{Vaswani_Attention_17} and graph residual terms \cite{zhang2019gresnet}, respectively. Here, we need to add some remarks on the graph residual terms used in the model. Different from \cite{zhang2020graph}, which integrates the target node residual terms to all the nodes in the batch, (1) for the graph instances without any node attributes, the nodes adjacency neighborhood vectors will be used to compute the residual terms, and (2) we compute the graph residual term with the whole graph instance instead in this paper (not merely for the target node).

Such a process will iterate through all the segments of the input graph instance nodes, and the finally learned node representations can be denoted as $\mb{H}^{(D)} = [\mb{h}_1^{(D)}, \mb{h}_2^{(D)}, \cdots, \mb{h}_{|\mc{V}|}^{(D)}]^\top \in \mathbbm{R}^{|\mc{V}| \times d_h}$. In this paper, for presentation simplicity, we just assume all the hidden layers are of the same dimension $d_h$. Considering that we focus on learning the representations of the entire graph instance in this paper, {\our} will integrate such node representations together to define the fused graph instance representation vector as follows:\vspace{-5pt}
\begin{equation}
\begin{aligned}
\mb{z}& = \mbox{Fusion} \left( \mb{H}^{(D)} \right) = \frac{1}{|\mc{V}|} \sum_{i=1}^{|\mc{V}|} \mb{h}_1^{(D)}.\vspace{-5pt}
\end{aligned}
\end{equation}
Both vector $\mb{z}$ and matrix $\mb{H}^{(D)}$ will be outputted to the down-stream application tasks for the model training/tuning and graph instance representation learning, which will be introduced in the follow-up section in detail.

\section{{\our} Learning}\label{sec:analysis}

In this part, we will focus on the pre-training and fine-tuning of {\our} with several concrete graph instance learning tasks. To be more specific, we will introduce two unsupervised pre-training tasks to learn {\our} based on \textit{node raw attribute reconstruction} and \textit{graph structure recovery}, which ensure the learned node and graph representations can capture both the raw attribute and structure information in the graph instances. After that, we will introduce one supervised tasks to fine-tune {\our} for the \textit{graph instance classification}, which will cast extra refinements on the learned node and graph representations.

\subsection{Unsupervised Pre-Training}

The pre-training tasks used in this paper will enable {\our} to effectively capture both the node raw attributes and graph instance structures in the learned nodes and graph representation vectors. In the case where the graph instances have no node raw attributes, only graph structure recovery will be used for the pre-training. Formally, based on the learned representations $\mb{H}^{(D)}$ of all the nodes in the graph instance, e.g., $G$, with several fully connected layers, we can project such latent representations to their raw features. Furthermore, by comparing the reconstructed feature matrix against the ground-truth raw features and minimizing their differences, we will be able to pre-train the {\our} model with all the graph instances. Furthermore, based on the nodes learned representations, we can also compute the closeness (e.g., cosine similarity of the representation vectors) of the node pairs in the graph. Furthermore, compared against the ground-truth graph connection weight matrix, we can define the introduced loss term for graph structure recovery of all the graph instances to pre-train the {\our} model.

\begin{table*}[t]
\caption{Experimental results of different comparison methods. For the results not reported in the previous works, we mark the corresponding entries with `$-$' in the table. The entries are the accuracy scores (mean$\pm$std) achieved by the baseline methods with the $10$ folds. For {\our}(padding/pruning, none) and {\our}(padding/pruning, raw), they denote {\our} with the \textit{padding/pruning} strategy and different graph residual terms (raw vs none). At the last row on {\our}*, we show the best results obtained by {\our} with all these three unification strategies.}\label{tab:result}
\vspace{-5pt}
\centering
\small
\setlength{\tabcolsep}{3pt}
\renewcommand{\arraystretch}{1.2}
\begin{tabular}{c|c|c|c|c|c|c|c}
\hline
\textbf{Datasets} &\textbf{IMDB-B} &\textbf{IMDB-M} &\textbf{COLLAB} &\textbf{MUTAG}  &\textbf{PROTEINS} &\textbf{PTC} &\textbf{NCI1}\\
\hline
\textbf{\# Graphs} &1,000 &1,500 &5,000 &188  &1,113 &344 &4,110 \\
\textbf{\# Classes} &2 &3 &3 &2  &2 &3 &2\\
\textbf{Avg. \# Nodes} &19.8 &13.0 &74.5 &17.9 &39.1 &25.5 &29.8\\
\textbf{max \# Nodes} &136 &89 &492 &28 &620 &109 &111\\
\hline
\hline
\textbf{Methods} & \multicolumn{7}{c}{\textbf{Accuracy} (mean$\pm$std)}\\
\hline

WL {\scriptsize \cite{Shervashidze_WL_11}} &73.40$\pm$4.63 &49.33$\pm$4.75 &\textbf{79.02$\pm$1.77} &82.05$\pm$0.36 &74.68$\pm$0.49 &59.90$\pm$4.30 &\textbf{82.19$\pm$0.18} \\

GK {\scriptsize \cite{pmlr-v5-shervashidze09a}} &65.87$\pm$0.98&43.89$\pm$0.38  &72.84$\pm$0.28&81.58$\pm$2.11&71.67$\pm$0.55&57.26$\pm$1.41&62.28$\pm$0.29 \\

\hline

DGK {\scriptsize \cite{Yanardag_Deep_15}} &66.96$\pm$0.56&44.55$\pm$0.52 &73.09$\pm$0.25&87.44$\pm$2.72&75.68$\pm$0.54&60.08$\pm$2.55&\textbf{80.31$\pm$0.46} \\

AWE {\scriptsize \cite{Ivanov_Anonymous_18}} &\textbf{74.45$\pm$5.83}&\textbf{51.54$\pm$3.61} &73.93$\pm$1.94&87.87$\pm$9.76&$-$&$-$&$-$ \\

PSCN {\scriptsize \cite{Niepert_Learning_16}} &71.00$\pm$2.29&45.23$\pm$2.84 &72.60$\pm$2.15&\textbf{88.95$\pm$4.37}&75.00$\pm$2.51&62.29$\pm$5.68&76.34$\pm$1.68 \\

\hline

DAGCN {\scriptsize \cite{Chen_Dual_19}}	&$-$&$-$&$-$&87.22$\pm$6.10	&76.33$\pm$4.3	&62.88$\pm$9.61	&\textbf{81.68$\pm$1.69}\\

SPI-GCN {\scriptsize \cite{SPIGCN}} &60.40$\pm$4.15	&44.13$\pm$4.61	&$-$	&84.40$\pm$8.14	&72.06$\pm$3.18	&56.41$\pm$5.71	&64.11$\pm$2.37\\

DGCNN {\scriptsize \cite{Zhang2018AnED}} &70.03$\pm$0.86&47.83$\pm$0.85 &73.76$\pm$0.49&85.83$\pm$1.66&75.54$\pm$0.94&58.59$\pm$2.47&74.44$\pm$0.47 \\

GCAPS-CNN {\scriptsize \cite{verma2018graph}} &71.69$\pm$3.40 &48.50$\pm$4.10 &77.71$\pm$2.51&$-$&\textbf{76.40$\pm$4.17}&\textbf{66.01$\pm$5.91}&\textbf{82.72$\pm$2.38} \\

CapsGNN {\scriptsize \cite{xinyi2018capsule}} &73.10$\pm$4.83 &50.27$\pm$2.65 &\textbf{79.62$\pm$0.91}&86.67$\pm$6.88&\textbf{76.28$\pm$3.63}&$-$&78.35$\pm$1.55 \\

\hline


{\our}(padding/pruning, none) &\textbf{75.40$\pm$2.29}&\textbf{52.27$\pm$1.55}&\textbf{78.42$\pm$1.29}&\textbf{89.24$\pm$7.78}&\textbf{77.09$\pm$4.15}&\textbf{68.86$\pm$4.17}&70.15$\pm$1.84 \\

{\our}(padding/pruning, raw) &\textbf{74.70$\pm$3.74}&\textbf{50.60$\pm$3.03}&74.90$\pm$1.78&\textbf{89.80$\pm$6.71}&\textbf{76.28$\pm$2.91}&\textbf{64.84$\pm$6.77}&68.10$\pm$2.55 \\

\hline

{\our}* &\textbf{77.20$\pm$3.09}&\textbf{53.40$\pm$2.12}&\textbf{78.42$\pm$1.29}&\textbf{90.85$\pm$6.58}&\textbf{77.09$\pm$4.15}&\textbf{68.86$\pm$4.17}&70.15$\pm$1.84\\

\hline
\end{tabular}
\vspace{-15pt}
\end{table*}

\subsection{Transfer and Fine-Tuning}

When applying the pre-trained {\our} and the learned representations in new application tasks, e.g., \textit{graph instance classification}, necessary fine-tuning can be needed. Formally, we can denote the batch of labeled graph instance as $\mc{T} = \left\{ (G_i, \mb{y}_i) \right\}_{i}$, where $\mb{y}_i$ denotes the label vector of graph instance $G_i$. Based on the fused representation vector $\mb{z}_i$ learned for graph instance $G_i \in \mc{G}$, we can further project it to the label vector with several fully connected layers together with the softmax normalization function, which can be denoted as
\begin{equation}
\hat{\mb{y}}_i = \mbox{softmax} \left( \mbox{FC} \left( \mb{z}_i \right) \right) \in \mathbbm{R}^{d_y \times 1}.
\end{equation}

Meanwhile, based on the known ground-truth label vector of graph instances in the training set, we can define the introduced loss term for the graph instance based on the corss-entropy term as follows:
\begin{equation}
\ell_{gc} = \sum_{(G_i, \mb{y}_i) \in \mc{T}} \sum_{j=1}^{d_y} - \mb{y}_i(j) \log \hat{\mb{y}}_i(j).
\end{equation}

%

By optimizing the above loss term, we will be able to re-fine {\our} and the learned graph instance representations based on the application tasks specifically.


\section{Experiments}\label{sec:experiment}

To test the effectiveness of {\our}, extensive experiments on real-world graph instance benchmark datasets will be done in this section. What's more, we will also compare {\our} with both the classic and state-of-the-art graph instance representation learning baseline methods to demonstrate its advantages. 

\noindent \textbf{Reproducibility}: Both the datasets and source code used can be accessed via the github page\footnote{https://github.com/jwzhanggy/SEG-BERT}. Detailed information about the server used to run the model can be found at the footnote\footnote{GPU Server: ASUS X99-E WS motherboard, Intel Core i7 CPU 6850K@3.6GHz (6 cores, 40 PCIe lanes), 3 Nvidia GeForce GTX 1080 Ti GPU (11 GB buffer each), 128 GB DDR4 memory and 128 GB SSD swap.}.

\begin{table*}[t]
\vspace{-7pt}
\caption{Evaluation results of different graph instance size unification strategies in {\our}. No residual terms are used here, and the default epoch number is $500$. The time denotes the average time cost for mode training in the 10 folds. For {\our} with the \textit{segment shifting} strategy, the default parameter $k$ is set as $20$.}\label{tab:strategy_result}
\vspace{-5pt}
\centering
\small
\setlength{\tabcolsep}{3.5pt}
\renewcommand{\arraystretch}{1.2}
\begin{tabular}{c|c|c|c||c|c|c||c|c|c||c|c|c}
\hline
\multirow{2}{*}{\textbf{Strategies}} &\multicolumn{3}{c||}{\textbf{IMDB-B}} &\multicolumn{3}{c||}{\textbf{IMDB-M}} &\multicolumn{3}{c||}{\textbf{MUTAG}} &\multicolumn{3}{c}{\textbf{PTC}}\\
\cline{2-13}
&Accuracy&k&Time(s)&Accuracy&k&Time(s)&Accuracy&k&Time(s)&Accuracy&k&Time(s)\\
\hline
\hline
Full-Input &\textbf{76.90$\pm$1.76}&136&2808.70&\textbf{53.33$\pm$2.53}&89&2397.42&\textbf{90.85$\pm$6.58}&28&55.73&68.01$\pm$4.23&109&700.98\\
\hline
Padding/Pruning 	&{75.40$\pm$2.29}	&50	&\textbf{1312.42}	&{52.27$\pm$1.55}	&50	&\textbf{1654.66}	&{89.24$\pm$7.78}	&25	&\textbf{55.19}	&\textbf{68.86$\pm$4.17}		&50	&\textbf{223.47}\\
\hline
Segment Shifting &\textbf{77.20$\pm$3.09}&20&1525.97&\textbf{53.40$\pm$2.12}&20&1730.08&90.29$\pm$7.74&20&88.40&66.54$\pm$4.18&20&295.83\\
\hline
\end{tabular}
\vspace{-10pt}
\end{table*}

\subsection{Dataset and Experimental Settings}

\noindent \textbf{Dataset Descriptions}: The graph instance datasets used in this experiments include IMDB-Binary, IMDB-Multi and COLLAB, as well as MUTAG, PROTEINS, PTC and NCI1, which are all the benchmark datasets as used in \cite{Yanardag_Deep_15} and all the follow-up graph classification papers \cite{xinyi2018capsule,verma2018graph,How_Xu_18}. Among them, IMDB-Binary, IMDB-Multi and COLLAB are the social graph datasets; whereas the remaining ones are the bio-graph datasets. Basic statistical information about the datasets is also available at the top of Table~\ref{tab:result}.


\noindent \textbf{Comparison Baselines}: The comparison baseline methods used in this paper include (1) \textbf{conventional graph kernel based methods}, e.g., Weisfeiler-Lehman subtree kernel (WL) \cite{Shervashidze_WL_11} and graphlet count kernel (GK) \cite{pmlr-v5-shervashidze09a}; (2) \textbf{existing deep learning based methods}, e.g., Deep Graph Kernel (DGK) \cite{Yanardag_Deep_15} and AWE \cite{Ivanov_Anonymous_18}, PATCHY-SAN (PSCN) \cite{Niepert_Learning_16}; and (3) \textbf{state-of-the-art deep learning methods}, e.g., Dual Attention Graph Convolutional Network (DAGCN) \cite{Chen_Dual_19}, Simple Permutation-Invariant Graph Convolutional Network (SPI-GCN) \cite{SPIGCN}, Graph Capsule CNN (GCAPS-CNN) \cite{verma2018graph} and Deep Graph CNN (DGCNN) \cite{Zhang2018AnED}, and Capsule Graph Neural Network (CapsGNN) \cite{xinyi2018capsule}. \textbf{Evaluation Metric}: The learning performance of these methods will be evaluated by Accuracy as the metric.

\noindent \textbf{Experimental Settings}: In the experiments, we will first compare {\our} with the \textit{padding/pruning} strategy for input size unification against the baseline methods, which can help test if a sub-structures in graph instances can capture the characteristics of the whole graph instance or not. The other two strategies will be discussed at the end of this section in detail. For the input portal size $k$, it is assigned with a value slightly larger than the graph instance average sizes of each dataset. All the graph instances in the datasets will be partitioned into train, validate, testing sets according to the ratio 8:1:1 with the $10$-fold cross validation, where the validation set is used for parameter tuning and selection. Meanwhile, for fair comparisons (existing works use $9:1$ train/test partition without validation set), such $10\%$ validation set will also be used as training data as well to further fine-tune {\our} after the parameter selection.

\noindent \textbf{Default Model Parameter Settings}: If not clearly specified, the results reported in this paper are based on the following parameter settings of {\our}: \textit{input portal size}: $k=25$ (MUTAG), $k=50$ (IMDB-Binary, IMDB-Multi, NCI1, PTC) and $k=100$ (COLLAB, PROTEINS); \textit{hidden size}: 32; \textit{attention head number}: 2; \textit{hidden layer number}: $D=2$; \textit{learning rate}: 0.0005 (PTC) and 0.0001 (others); \textit{weight decay}: $5e^{-4}$; \textit{intermediate size}: 32; \textit{hidden dropout rate}: 0.5; \textit{attention dropout rate}: 0.3; \textit{graph residual term}: raw/none; \textit{training epoch}: 500 (early stop if necessary to avoid over-fitting).


\subsection{Graph Instance Classification Results}\label{subsec:main_result}

\noindent \textbf{Model Train Records}: As illustrated in Figure~\ref{fig:graph_bert_protein}, we show the learning records (training/testing accuracy) of {\our} on both the IMDB-Multi social graph and the Protein bio-graph datasets. According to the plots, graph instance classification is very different from classification task on other data types, as the model can get over-fitting easily. Similar phenomena have been observed for most comparison methods on the other datasets as well. Even though the default training epoch is $500$ mentioned above, an early stop of training {\our} is usually necessary. In this paper, we decide the early-stop learning epoch number based on the partitioned validation set.

\noindent \textbf{Main Results}: The main learning results of of all the comparison methods are provided in Table~\ref{tab:result}. The {\our} method shown here adopts the \textit{padding/pruning} strategy to handle the graph data input. The residual terms adopted are indicated by the {\our} method name in the parentheses. According to the evaluation results, {\our} can greatly improve the learning performance on most of the benchmark datasets. For instance, the accuracy score of {\our} (none) on IMDB-Binary is $75.40$, which is much higher than many of the state-of-the-art baseline methods, e.g., DGCNN, GCAPS-CNN and CapsGNN. Similarly learning performance advantages have also been observed on the other datasets, except NCI1. 

For the raw residual term used in {\our}, for some of the datasets as shown in Table~\ref{tab:result}, e.g., MUTAG, it can improve the learning performance; whereas for the remaining ones, its effectiveness is not very significant. The main reason can be for most of the graph instances studied here they don't have node attributes and the discrete nodes adjacency neighborhood embeddings can be very sparse, which renders the computed residual terms to be less effective for performance improvement.

\noindent \textbf{Graph Size Unification Strategy Analysis}: The analyses of the different graph size unification strategies is provided in Table~\ref{tab:strategy_result}. For {\our} with the \textit{full-input strategy} on COLLAB, PROTEIN and NCI1 (with more instances and large max graph sizes), the training time costs are too high ($>$ 3 days to run the 10 folds). So, the results on these three datasets are not shown here. As illustrated in Table~\ref{tab:strategy_result}, the performance of \textit{full-input} is better than \textit{padding/pruning}, but it also consumes the highest time cost. The \textit{padding/pruning} strategy has the lowest time costs but its performance is slightly lower than \textit{full-input} and \textit{segment shifting}. Meanwhile, the learning performance of \textit{segment shifting} balances between \textit{full-input} and \textit{padding/pruning}, which can even out-perform \textit{full-input} as it introduce far less dummy paddings in the input.

\section{Conclusion}\label{sec:conclusion}

In this paper, we have introduce a new graph neural network model, namely {\our}, for graph instance representation learning. With several significant modifications, the re-designed {\our} has no node-order-variant inputs or function components, which can handle the node orderless property very well. {\our} has an extendable architecture. For large-sized graph instance input, {\our} will divide the nodes into segments, whereas all the nodes in the graph will be used for the representation learning of each segment. We have tested the effectiveness of {\our} on several concrete application tasks, and the experimental results demonstrate that {\our} can out-perform the state-of-the-art graph instance representation learning models effectively.
\bibliography{reference}

\begin{thebibliography}{37}
\providecommand{\natexlab}[1]{#1}
\providecommand{\url}[1]{\texttt{#1}}
\expandafter\ifx\csname urlstyle\endcsname\relax
  \providecommand{\doi}[1]{doi: #1}\else
  \providecommand{\doi}{doi: \begingroup \urlstyle{rm}\Url}\fi

\bibitem[Atamna et~al.(2019)Atamna, Sokolovska, and CRIVELLO]{SPIGCN}
Atamna, A., Sokolovska, N., and CRIVELLO, J.-C.
\newblock {SPI-GCN: A Simple Permutation-Invariant Graph Convolutional
  Network}.
\newblock working paper or preprint, April 2019.
\newblock URL \url{https://hal.archives-ouvertes.fr/hal-02093451}.

\bibitem[Bondy(1976)]{Bondy_Graph_76}
Bondy, J.~A.
\newblock \emph{Graph Theory With Applications}.
\newblock Elsevier Science Ltd., GBR, 1976.
\newblock ISBN 0444194517.

\bibitem[Chen et~al.(2019)Chen, Pan, Jiang, Huo, and Long]{Chen_Dual_19}
Chen, F., Pan, S., Jiang, J., Huo, H., and Long, G.
\newblock {DAGCN:} dual attention graph convolutional networks.
\newblock \emph{CoRR}, abs/1904.02278, 2019.
\newblock URL \url{http://arxiv.org/abs/1904.02278}.

\bibitem[Chung et~al.(2003)Chung, Lu, and Vu]{Chung_Spectra_03}
Chung, F., Lu, L., and Vu, V.
\newblock Spectra of random graphs with given expected degrees.
\newblock \emph{Proceedings of the National Academy of Sciences}, 100\penalty0
  (11):\penalty0 6313--6318, 2003.
\newblock ISSN 0027-8424.
\newblock \doi{10.1073/pnas.0937490100}.
\newblock URL \url{https://www.pnas.org/content/100/11/6313}.

\bibitem[Chung et~al.(2014)Chung, G{\"{u}}l{\c{c}}ehre, Cho, and
  Bengio]{DBLP:journals/corr/ChungGCB14}
Chung, J., G{\"{u}}l{\c{c}}ehre, {\c{C}}., Cho, K., and Bengio, Y.
\newblock Empirical evaluation of gated recurrent neural networks on sequence
  modeling.
\newblock \emph{CoRR}, abs/1412.3555, 2014.
\newblock URL \url{http://arxiv.org/abs/1412.3555}.

\bibitem[Defferrard et~al.(2016)Defferrard, Bresson, and
  Vandergheynst]{NIPS2016_6081}
Defferrard, M., Bresson, X., and Vandergheynst, P.
\newblock Convolutional neural networks on graphs with fast localized spectral
  filtering.
\newblock In Lee, D.~D., Sugiyama, M., Luxburg, U.~V., Guyon, I., and Garnett,
  R. (eds.), \emph{Advances in Neural Information Processing Systems 29}, pp.\
  3844--3852. Curran Associates, Inc., 2016.

\bibitem[Devlin et~al.(2018)Devlin, Chang, Lee, and Toutanova]{Bert}
Devlin, J., Chang, M., Lee, K., and Toutanova, K.
\newblock {BERT:} pre-training of deep bidirectional transformers for language
  understanding.
\newblock \emph{CoRR}, abs/1810.04805, 2018.
\newblock URL \url{http://arxiv.org/abs/1810.04805}.

\bibitem[Hammond et~al.(2011)Hammond, Vandergheynst, and
  Gribonval]{Hammond_2011}
Hammond, D.~K., Vandergheynst, P., and Gribonval, R.
\newblock Wavelets on graphs via spectral graph theory.
\newblock \emph{Applied and Computational Harmonic Analysis}, 30\penalty0
  (2):\penalty0 129?150, Mar 2011.
\newblock ISSN 1063-5203.
\newblock \doi{10.1016/j.acha.2010.04.005}.
\newblock URL \url{http://dx.doi.org/10.1016/j.acha.2010.04.005}.

\bibitem[Hochreiter \& Schmidhuber(1997)Hochreiter and
  Schmidhuber]{Hochreiter_Long_Neural_97}
Hochreiter, S. and Schmidhuber, J.
\newblock Long short-term memory.
\newblock \emph{Neural Comput.}, 9\penalty0 (8), November 1997.

\bibitem[Ivanov \& Burnaev(2018)Ivanov and Burnaev]{Ivanov_Anonymous_18}
Ivanov, S. and Burnaev, E.
\newblock Anonymous walk embeddings.
\newblock \emph{CoRR}, abs/1805.11921, 2018.
\newblock URL \url{http://arxiv.org/abs/1805.11921}.

\bibitem[Jiang et~al.(2018)Jiang, Cui, Xu, and Yang]{Jiang_Gaussian_18}
Jiang, J., Cui, Z., Xu, C., and Yang, J.
\newblock Gaussian-induced convolution for graphs.
\newblock \emph{CoRR}, abs/1811.04393, 2018.
\newblock URL \url{http://arxiv.org/abs/1811.04393}.

\bibitem[Kim(2014)]{kim-2014-convolutional}
Kim, Y.
\newblock Convolutional neural networks for sentence classification.
\newblock In \emph{Proceedings of the 2014 Conference on Empirical Methods in
  Natural Language Processing ({EMNLP})}, pp.\  1746--1751, Doha, Qatar,
  October 2014. Association for Computational Linguistics.
\newblock \doi{10.3115/v1/D14-1181}.
\newblock URL \url{https://www.aclweb.org/anthology/D14-1181}.

\bibitem[Kipf \& Welling(2016)Kipf and Welling]{Kipf_Semi_CORR_16}
Kipf, T.~N. and Welling, M.
\newblock Semi-supervised classification with graph convolutional networks.
\newblock \emph{CoRR}, abs/1609.02907, 2016.

\bibitem[Kriege et~al.(2016)Kriege, Giscard, and Wilson]{NIPS2016_6166}
Kriege, N.~M., Giscard, P.-L., and Wilson, R.
\newblock On valid optimal assignment kernels and applications to graph
  classification.
\newblock In Lee, D.~D., Sugiyama, M., Luxburg, U.~V., Guyon, I., and Garnett,
  R. (eds.), \emph{Advances in Neural Information Processing Systems 29}, pp.\
  1623--1631. Curran Associates, Inc., 2016.

\bibitem[Krizhevsky et~al.(2012)Krizhevsky, Sutskever, and
  Hinton]{NIPS2012_4824}
Krizhevsky, A., Sutskever, I., and Hinton, G.~E.
\newblock Imagenet classification with deep convolutional neural networks.
\newblock In Pereira, F., Burges, C. J.~C., Bottou, L., and Weinberger, K.~Q.
  (eds.), \emph{Advances in Neural Information Processing Systems 25}, pp.\
  1097--1105. Curran Associates, Inc., 2012.

\bibitem[Liu et~al.(2019)Liu, Ott, Goyal, Du, Joshi, Chen, Levy, Lewis,
  Zettlemoyer, and Stoyanov]{Liu_RoBERTa}
Liu, Y., Ott, M., Goyal, N., Du, J., Joshi, M., Chen, D., Levy, O., Lewis, M.,
  Zettlemoyer, L., and Stoyanov, V.
\newblock Roberta: {A} robustly optimized {BERT} pretraining approach.
\newblock \emph{CoRR}, abs/1907.11692, 2019.
\newblock URL \url{http://arxiv.org/abs/1907.11692}.

\bibitem[Lov\'asz(1996)]{Lovasz1996}
Lov\'asz, L.
\newblock Random walks on graphs: A survey.
\newblock In {Mikl\'os}, D., {S\'os}, V.~T., and {Sz\H{o}nyi}, T. (eds.),
  \emph{Combinatorics, Paul Erd\H{o}s is Eighty}, volume~2, pp.\  353--398.
  J\'anos Bolyai Mathematical Society, Budapest, 1996.

\bibitem[Mallea et~al.(2019)Mallea, Meltzer, and Bentley]{Mallea_Capsule_19}
Mallea, M. D.~G., Meltzer, P., and Bentley, P.~J.
\newblock Capsule neural networks for graph classification using explicit
  tensorial graph representations.
\newblock \emph{CoRR}, abs/1902.08399, 2019.
\newblock URL \url{http://arxiv.org/abs/1902.08399}.

\bibitem[Meltzer et~al.(2019)Meltzer, Mallea, and
  Bentley]{Meltzer_Permutation_19}
Meltzer, P., Mallea, M. D.~G., and Bentley, P.~J.
\newblock Pinet: {A} permutation invariant graph neural network for graph
  classification.
\newblock \emph{CoRR}, abs/1905.03046, 2019.
\newblock URL \url{http://arxiv.org/abs/1905.03046}.

\bibitem[Meng \& Zhang(2019)Meng and Zhang]{Meng_Isomorphic_NIPS_19}
Meng, L. and Zhang, J.
\newblock Isonn: Isomorphic neural network for graph representation learning
  and classification.
\newblock \emph{CoRR}, abs/1907.09495, 2019.
\newblock URL \url{http://arxiv.org/abs/1907.09495}.

\bibitem[Narayanan et~al.(2017)Narayanan, Chandramohan, Venkatesan, Chen, Liu,
  and Jaiswal]{Narayanan_Graph_17}
Narayanan, A., Chandramohan, M., Venkatesan, R., Chen, L., Liu, Y., and
  Jaiswal, S.
\newblock graph2vec: Learning distributed representations of graphs.
\newblock \emph{CoRR}, abs/1707.05005, 2017.
\newblock URL \url{http://arxiv.org/abs/1707.05005}.

\bibitem[Niepert et~al.(2016)Niepert, Ahmed, and Kutzkov]{Niepert_Learning_16}
Niepert, M., Ahmed, M., and Kutzkov, K.
\newblock Learning convolutional neural networks for graphs.
\newblock \emph{CoRR}, abs/1605.05273, 2016.
\newblock URL \url{http://arxiv.org/abs/1605.05273}.

\bibitem[Raffel et~al.(2019)Raffel, Shazeer, Roberts, Lee, Narang, Matena,
  Zhou, Li, and Liu]{raffel2019exploring}
Raffel, C., Shazeer, N., Roberts, A., Lee, K., Narang, S., Matena, M., Zhou,
  Y., Li, W., and Liu, P.~J.
\newblock Exploring the limits of transfer learning with a unified text-to-text
  transformer.
\newblock \emph{arXiv preprint arXiv:1910.10683}, 2019.

\bibitem[Ranjan et~al.(2019)Ranjan, Sanyal, and Talukdar]{ranjan2019asap}
Ranjan, E., Sanyal, S., and Talukdar, P.~P.
\newblock Asap: Adaptive structure aware pooling for learning hierarchical
  graph representations.
\newblock \emph{arXiv preprint arXiv:1911.07979}, 2019.

\bibitem[Shervashidze et~al.(2009)Shervashidze, Vishwanathan, Petri, Mehlhorn,
  and Borgwardt]{pmlr-v5-shervashidze09a}
Shervashidze, N., Vishwanathan, S., Petri, T., Mehlhorn, K., and Borgwardt, K.
\newblock Efficient graphlet kernels for large graph comparison.
\newblock In van Dyk, D. and Welling, M. (eds.), \emph{Proceedings of the
  Twelth International Conference on Artificial Intelligence and Statistics},
  volume~5 of \emph{Proceedings of Machine Learning Research}, pp.\  488--495,
  Hilton Clearwater Beach Resort, Clearwater Beach, Florida USA, 16--18 Apr
  2009. PMLR.
\newblock URL \url{http://proceedings.mlr.press/v5/shervashidze09a.html}.

\bibitem[Shervashidze et~al.(2011)Shervashidze, Schweitzer, van Leeuwen,
  Mehlhorn, and Borgwardt]{Shervashidze_WL_11}
Shervashidze, N., Schweitzer, P., van Leeuwen, E.~J., Mehlhorn, K., and
  Borgwardt, K.~M.
\newblock Weisfeiler-lehman graph kernels.
\newblock \emph{J. Mach. Learn. Res.}, 12\penalty0 (null):\penalty0 2539?2561,
  November 2011.
\newblock ISSN 1532-4435.

\bibitem[Sun et~al.(2019)Sun, Wang, Li, Feng, Tian, Wu, and Wang]{Sun_ERNIE}
Sun, Y., Wang, S., Li, Y., Feng, S., Tian, H., Wu, H., and Wang, H.
\newblock {ERNIE} 2.0: {A} continual pre-training framework for language
  understanding.
\newblock \emph{CoRR}, abs/1907.12412, 2019.
\newblock URL \url{http://arxiv.org/abs/1907.12412}.

\bibitem[Vaswani et~al.(2017)Vaswani, Shazeer, Parmar, Uszkoreit, Jones, Gomez,
  Kaiser, and Polosukhin]{Vaswani_Attention_17}
Vaswani, A., Shazeer, N., Parmar, N., Uszkoreit, J., Jones, L., Gomez, A.~N.,
  Kaiser, L., and Polosukhin, I.
\newblock Attention is all you need.
\newblock \emph{CoRR}, abs/1706.03762, 2017.
\newblock URL \url{http://arxiv.org/abs/1706.03762}.

\bibitem[Veli{\v{c}}kovi{\'{c}} et~al.(2018)Veli{\v{c}}kovi{\'{c}}, Cucurull,
  Casanova, Romero, Li{\`{o}}, and Bengio]{Velickovic_Graph_ICLR_18}
Veli{\v{c}}kovi{\'{c}}, P., Cucurull, G., Casanova, A., Romero, A., Li{\`{o}},
  P., and Bengio, Y.
\newblock {Graph Attention Networks}.
\newblock \emph{International Conference on Learning Representations}, 2018.

\bibitem[Verma \& Zhang(2018)Verma and Zhang]{verma2018graph}
Verma, S. and Zhang, Z.-L.
\newblock Graph capsule convolutional neural networks.
\newblock \emph{arXiv preprint arXiv:1805.08090}, 2018.

\bibitem[Xu et~al.(2018)Xu, Hu, Leskovec, and Jegelka]{How_Xu_18}
Xu, K., Hu, W., Leskovec, J., and Jegelka, S.
\newblock How powerful are graph neural networks?
\newblock \emph{CoRR}, abs/1810.00826, 2018.
\newblock URL \url{http://arxiv.org/abs/1810.00826}.

\bibitem[Yanardag \& Vishwanathan(2015)Yanardag and
  Vishwanathan]{Yanardag_Deep_15}
Yanardag, P. and Vishwanathan, S.
\newblock Deep graph kernels.
\newblock In \emph{Proceedings of the 21th ACM SIGKDD International Conference
  on Knowledge Discovery and Data Mining}, KDD ?15, pp.\  1365?1374, New York,
  NY, USA, 2015. Association for Computing Machinery.
\newblock ISBN 9781450336642.
\newblock \doi{10.1145/2783258.2783417}.
\newblock URL \url{https://doi.org/10.1145/2783258.2783417}.

\bibitem[{Zhang}(2019)]{zhang2019graph}
{Zhang}, J.
\newblock {Graph Neural Networks for Small Graph and Giant Network
  Representation Learning: An Overview}.
\newblock \emph{arXiv e-prints}, art. arXiv:1908.00187, Jul 2019.

\bibitem[Zhang \& Meng(2019)Zhang and Meng]{zhang2019gresnet}
Zhang, J. and Meng, L.
\newblock Gresnet: Graph residual network for reviving deep gnns from suspended
  animation.
\newblock \emph{ArXiv, abs/1909.05729}, 2019.

\bibitem[Zhang et~al.(2020)Zhang, Zhang, Sun, and Xia]{zhang2020graph}
Zhang, J., Zhang, H., Sun, L., and Xia, C.
\newblock Graph-bert: Only attention is needed for learning graph
  representations.
\newblock \emph{arXiv preprint arXiv:2001.05140}, 2020.

\bibitem[Zhang et~al.(2018)Zhang, Cui, Neumann, and Chen]{Zhang2018AnED}
Zhang, M., Cui, Z., Neumann, M., and Chen, Y.
\newblock An end-to-end deep learning architecture for graph classification.
\newblock In \emph{AAAI}, 2018.

\bibitem[Zhang \& Chen(2019)Zhang and Chen]{xinyi2018capsule}
Zhang, X. and Chen, L.
\newblock Capsule graph neural network.
\newblock In \emph{International Conference on Learning Representations}, 2019.
\newblock URL \url{https://openreview.net/forum?id=Byl8BnRcYm}.

\end{thebibliography}
\bibliographystyle{icml2020}


\end{document}